\documentclass[lettersize,journal]{IEEEtran}
\usepackage{amsmath,amsfonts}
\usepackage{algorithmic}
\usepackage{algorithm}
\usepackage{array}
\usepackage[caption=false,font=footnotesize,labelfont=rm,textfont=rm]{subfig}
\usepackage{textcomp}
\usepackage{stfloats}
\usepackage{url}
\usepackage{verbatim}
\usepackage{graphicx}
\usepackage{cite}
\usepackage{multirow}
\usepackage{stfloats}
\usepackage{booktabs}
\usepackage{color}
\usepackage{flushend}
\usepackage{balance} 
\makeatletter
\renewcommand{\maketag@@@}[1]{\hbox{\m@th\normalsize\normalfont#1}}%
\makeatother
\begin{document}

\title{Dual Contrastive Network for Few-Shot Remote Sensing Image Scene Classification}

\author{Zhong Ji,~\IEEEmembership{Senior Member,~IEEE,} Liyuan Hou, Xuan Wang, Gang Wang, 

and Yanwei Pang,~\IEEEmembership{Senior Member,~IEEE}
\thanks{This work was supported by the National Natural Science Foundation of China under Grant 62176178. \emph{(Corresponding author: Xuan Wang.)}

Zhong Ji is with the School of Electrical and Information Engineering, Tianjin University, Tianjin 300072, China, and also with the CETC Key Laboratory of Aerospace Information Applications, Shijiazhuang, 050081, China.

Liyuan Hou, Xuan Wang, and Yanwei Pang are with the School of Electrical and Information Engineering, Tianjin University, Tianjin 300072, China (e-mail: wang\_xuan@tju.edu.cn).

Gang Wang is with the CETC Key Laboratory of Aerospace Information Applications, Shijiazhuang, 050081, China.}}

\markboth{Journal of \LaTeX\ Class Files,~Vol.~14, No.~8, August~2021}%
{Shell \MakeLowercase{\textit{et al.}}: A Sample Article Using IEEEtran.cls for IEEE Journals}

\IEEEpubid{0000--0000/00\$00.00~\copyright~2021 IEEE}

\maketitle

\begin{abstract}
Few-shot remote sensing image scene classification (FS-RSISC) aims at classifying remote sensing images with only a few labeled samples. The main challenges lie in small inter-class variances and large intra-class variances, which are the inherent property of remote sensing images. To address these challenges, we propose a transfer-based Dual Contrastive Network (DCN), which incorporates two auxiliary supervised contrastive learning branches during the training process. Specifically, one is a Context-guided Contrastive Learning (CCL) branch and the other is a Detail-guided Contrastive Learning (DCL) branch, which focus on inter-class discriminability and intra-class invariance, respectively. In the CCL branch, we first devise a Condenser Network to capture context features, and then leverage a supervised contrastive learning on top of the obtained context features to facilitate the model to learn more discriminative features. In the DCL branch, a Smelter Network is designed to highlight the significant local detail information. And then we construct a supervised contrastive learning based on the detail feature maps to fully exploit the spatial information in each map, enabling the model to concentrate on invariant detail features. Extensive experiments on four public benchmark remote sensing datasets demonstrate the competitive performance of our proposed DCN. 
\end{abstract}


\section{Introduction}
\IEEEPARstart{R}{emote} sensing image scene classification (RSISC) aims at predicting the scene labels of remote sensing images, which is widely applied in the various fields, such as disaster detection \cite{15,54,68}, environmental monitoring \cite{16,55}, urban planning \cite{17,52}, and land-use management \cite{53,69}. In these fields, deep learning \cite{19,20,66,67,83} has recently been leveraged to perform the RSISC task with advanced results. However, these deep learning approaches are data-driven, which suffer from the lack of labeled remote sensing images in real world. To alleviate this limitation, the Few-Shot RSISC (FS-RSISC) \cite{9,10} is proposed to address the RSISC with only a few labeled samples.

Most existing FS-RSISC approaches fall in the group of meta-learning \cite{9,10,11,12}. These meta-based approaches train a meta-learner with a diverse set of few-shot tasks, which transfers the knowledge to novel tasks, and have achieved remarkable success in FS-RSISC. However, they underestimate the fact that a good feature extractor is still fundamental for few-shot learning \cite{41,14}, which is confirmed in recent researches \cite{45,46}. Inspired by this evidence, a transfer-based approach \cite{2} was proposed for FS-RSISC by designing a mirror-based feature extractor to construct sub-supervised data and corresponding labels to assist the pre-training stage.

\IEEEpubidadjcol

\begin{figure}[t]
\centering
\includegraphics[scale=0.4]{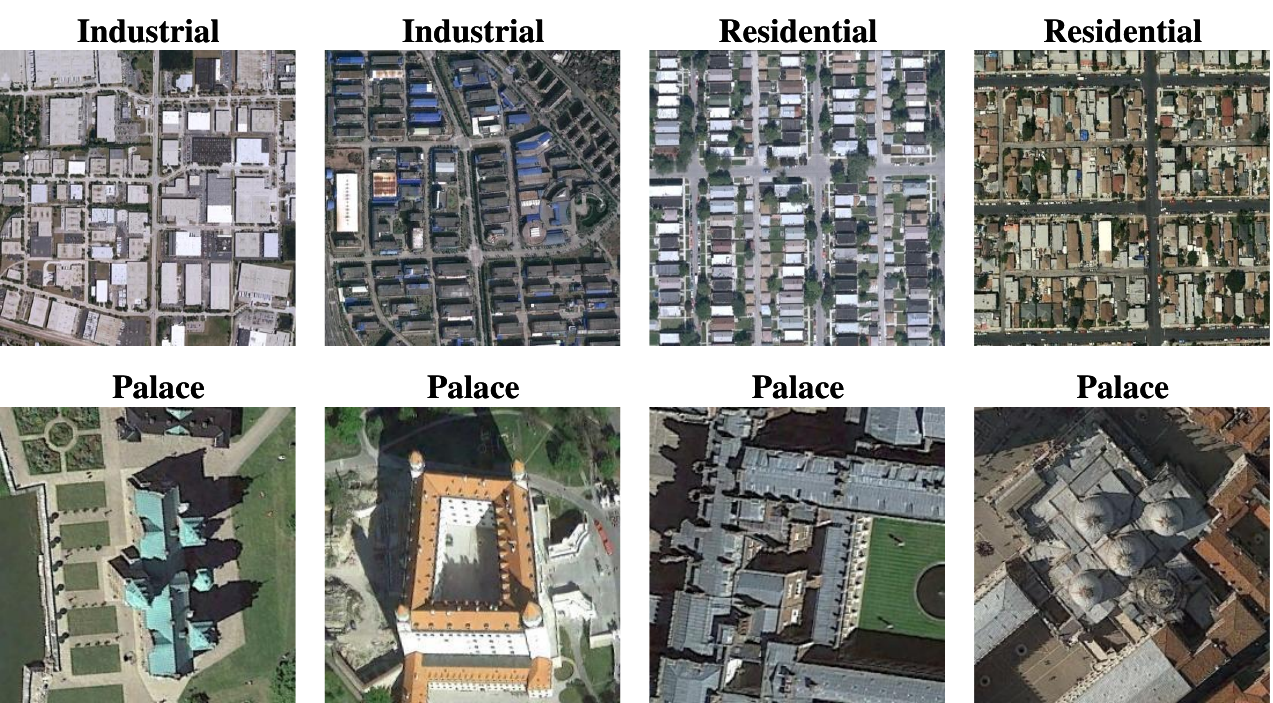}
\caption{Illustrations of remote sensing image characteristics: small inter-class variances and large intra-class variances. See texts for details.}\label{fig1}
\end{figure}

Nevertheless, \cite{2} fails to train a generalized feature extractor due to neglecting the inherent property of scene images: small inter-class variances and large intra-class variances. For example, as shown in Fig.\ref{fig1}, residential and industrial classes in the first row are highly similar in terms of the objects and their arrangements, which result in a small inter-class variance. While the second row exhibits four remote sensing images belonging to the same palace scene, in which the contents vary from backgrounds to appearances. This dissimilarity of the same class images leads to a large intra-class variance. Although some existing approaches have addressed the challenges of RSISC with feature fusion and attention mechanism  \cite{63,64,65}, they left the complementary relationship between context and detail information out of consideration.

To this end, we present a Dual Contrastive Network (DCN) containing two supervised contrastive learning branches to pull features from the same class closer than those from different classes, which facilitates the model to explore the context and detail features and thus improves the model generalization, as illustrated in Fig. \ref{fig2}. Specifically, to alleviate the problem of small inter-class variances, we propose a Context-guided Contrastive Learning (CCL) branch, which aims at improving the discrimination of each category to enable the model to distinguish inter-class differences. Particularly, we first devise a Condenser Network to condense the global context information by integrating and redistributing the information at each spatial location via the Squeeze and Expand operations among channels. It imitates the condensation process and enhances the information relevant to the scene while ignoring the irrelevant information, which results in more representative context features. Then, the context features are employed for a contrastive learning, which is an auxiliary task for scene classification in the training process. The CCL branch enables the model to aggregate vital information from all channels, thus promoting the discriminative feature extraction capability of the model.

To address the problem of large intra-class variances, we further develop a Detail-guided Contrastive Learning (DCL) branch based on feature maps, with the goal of capturing the detail features that are invariant in intra-class samples by leveraging the spatial information contained in the feature maps. Similar to the process of smelting, we first devise a Smelter Network to adaptively locate the significant local regions of the input images and smelt the corresponding detail features. To be specific, the locally detailed feature information in both channel and spatial dimensions is emphasized. Then the channel and spatial information are aggregated together and mapped into attention weights, which are applied to enhance the information on significant local regions of the original feature maps. The enhanced detail feature maps are employed to construct an auxiliary contrastive learning task, which facilitates the model to concentrate on the spatial detail information of the feature maps. Notably, we introduce a self-attention based feature alignment mechanism to accurately measure the similarity between two samples.

In summary, the main highlights are four-fold:

\begin{enumerate}[]
\item We propose a Dual Contrastive Network (DCN) for FS-RSISC, which consists of a Context-guided Contrastive Learning (CCL) and a Detail-guided Contrastive Learning (DCL). It enables the model to extract discriminative and detail features by the complementarity between the two contrastive learning branches.

\item We design a Condenser Network for CCL, which enhances the information relevant to the scene by exploiting the global contexts to generate discriminative context features.

\item A Smelter Network is devised for DCL, which extracts detail features by adaptively localizing the significant local image regions.

\item Our DCN achieves competitive performance on four benchmark remote sensing datasets, which even outperforms state-of-the-art approaches by at least 3.56$\%$ and 5.38$\%$ on the NWPU-RESISC45 \cite{24} and AID \cite{27} datasets on 5-way 5-shot, respectively. 
\end{enumerate}

The rest of the paper is organized as follows: Section II introduces the related work. The overall architecture of the proposed DCN is described in Section III. Section IV shows experimental results on four benchmark remote sensing datasets. Finally, the conclusion is summarized in Section V.

\section{Related Work}
\subsection{Remote Sensing Image Scene Classification}
With the renascence of neural network, RSISC has received great attention \cite{34,35,57,84,37,38,39}. For example, Han \emph{et al}. \cite{35} devised an improved pre-trained AlexNet architecture for scene classification, which incorporated the spatial pyramid pooling and side supervision to fuse multi-scale information of remote sensing images. He \emph{et al}. \cite{57} proposed an end-to-end learning model called skip-connected covariance network to achieve more representative feature learning. Zhang \emph{et al}. \cite{37} made full use of the merits of convolutional neural network (CNN) and CapsNet to improve the classification performance. Liu \emph{et al}. \cite{38} designed a Siamese CNN to mitigate the impact of the small scale of scene classes and lack of image diversity. Zeng \emph{et al}. \cite{39} proposed a prototype calibration with a feature-generating model to solve the deviation of prototype feature expression. 

In addition, transfer learning with large-scale pre-training CNN is also widely exploited in RSISC \cite{74,75,76,77,78}. For instance, Lima \emph{et al}. \cite{74} proved that it is possible to transfer from natural images (ImageNet) to remote-sensing imagery by using pre-trained VGG19 and Inception V3 on remote sensing datasets. Xie \emph{et al}. \cite{75} introduced a scale-free CNN by converting the fully connected layers in the pre-trained CNN model to convolutional layers and using a global average pooling layer after the final convolutional layer, which addressed the information losing issue during the fine-tuning process. Li \emph{et al}. \cite{76} trained three typical deep CNN models and then transferred the weights of their convolutional layers to prevent overfitting. Wang \emph{et al}. \cite{77} proposed an adaptive learning strategy to manage the input, model, and label simultaneously for transferring a CNN-based model. Alem \emph{et al}. \cite{78} developed a novel deep transfer learning-based fusion model which combined multiple feature vectors to resolve the semantic gap among various datasets.

Recently, with the diversity of remote sensing images and fine division of scene categories, some studies have begun to address the problems of small inter-class variances and large intra-class variances of remote sensing images \cite{49,47,50,51,71}. Yu \emph{et al}. \cite{49} investigated texture coded and saliency coded two-stream deep architectures to achieve the feature-level fusion. Goel \emph{et al}. \cite{47} adopted a hierarchical metric learning strategy, which organized the classes in a hierarchical fashion to explore the class interaction information. Shen \emph{et al}. \cite{50} constructed a dual-model architecture with a grouping-attention-fusion strategy to enhance the feature representation of the scene. Chen \emph{et al}. \cite{51} presented a Multi-Branch Local Attention Network, which emphasized the main target in the complex background by both channel and spatial attentions. However, these studies only concentrate on extracting critical local features and fuse them with the original features. Unlike them, our work not only extracts local details, but also explores the global context information, which achieves the complementarity between the obtained detail feature maps and context features.
\subsection{Few-shot Remote Sensing Image Scene Classification}
Few-shot learning aims at classifying samples from unseen novel classes with a few labeled samples, which has achieved extensive progress \cite{6,7,8,5,29} and has recently been applied to RSISC \cite{70,9,10,62,1,11}. For example, Zhai \emph{et al}. \cite{9} presented a lifelong few-shot learning model based on meta-learning with gradient descent, which enabled the learned knowledge to be easily and rapidly adapted to a new dataset and in turn served in the lifelong few-shot learning. Li \emph{et al}. \cite{10} proposed a discriminative learning of adaptive match network (DLA-MatchNet) including a feature learning module and matcher to learn more discriminative feature representations and a robust metric. Kim \emph{et al}. \cite{62} devised a multi-scale feature fusion network by integrating a novel self-attention feature selection module to extract more valuable texture semantic features from a limited number of labeled input images. Cheng \emph{et al}. \cite{1} introduced a Siamese-prototype network (SPNet) with prototype self-calibration and inter-calibration to learn more valid and representative prototypes. Li \emph{et al}. \cite{11} adopted a self-supervised contrastive learning-based metric learning network (SCL-MLNet) by weaving self-supervised contrastive learning into few-shot classification algorithms to promote the scene classification performance. Although great advance has been made, most of the existing studies are based on meta-learning framework and underestimate the importance of training a good feature extractor. Our work falls in the group of transfer learning which aims at learning a robust and generalized feature representation. Moreover, different from the above methods with a single classification model, our approach contains two branches that focus on discriminative features and detail features respectively, to adapt to scene classification tasks of different complexity.

\begin{figure*}[t]
\centering
\includegraphics[scale=0.48]{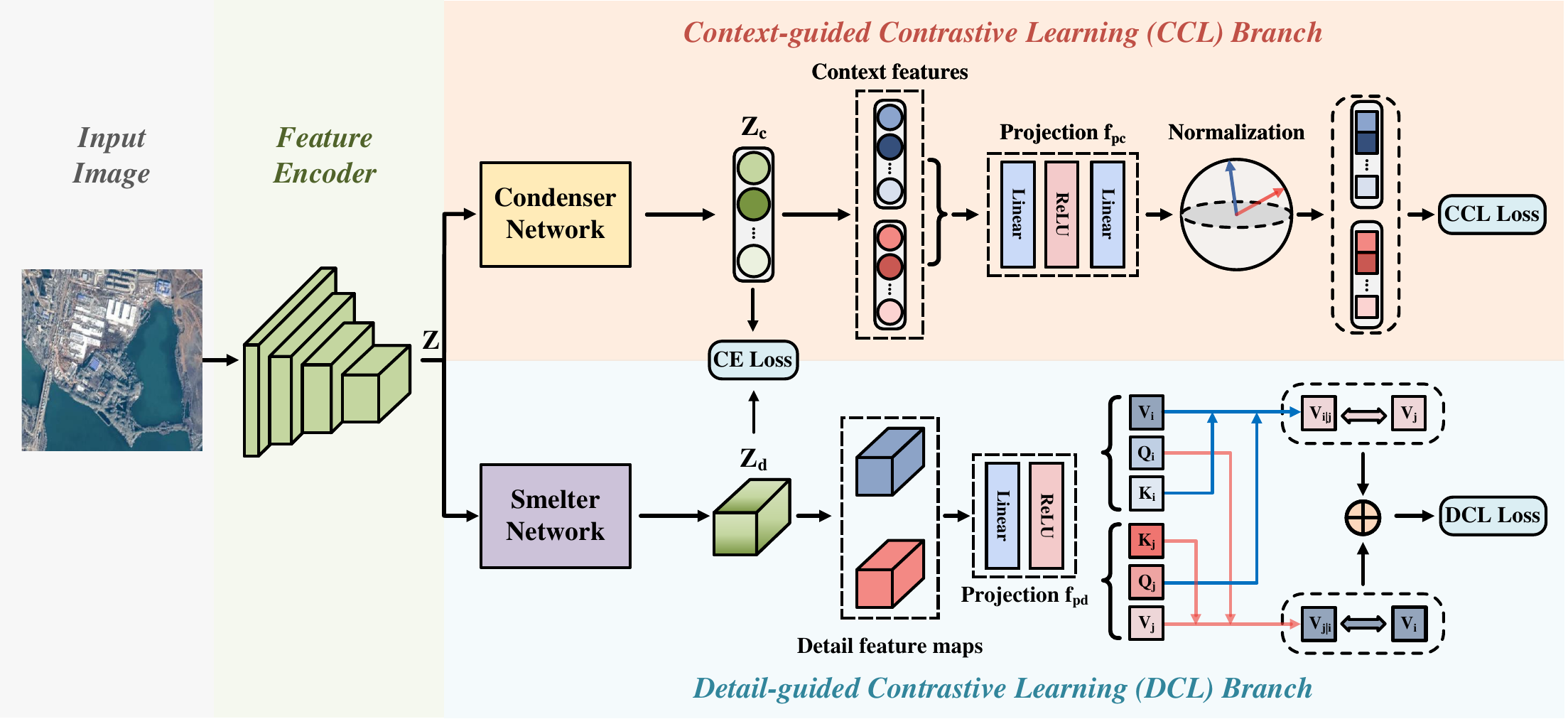}
\caption{Overview of the proposed DCN framework. It consists of a Context-guided Contrastive Learning (CCL) branch and a Detail-guided Contrastive Learning (DCL) branch to facilitate the model to learn more discriminative and detail information to generate context and detail features, both of which are used for training their corresponding classifiers.}\label{fig2}
\end{figure*}

\section{Method}
In this section, we first describe the FS-RSISC problem, and then introduce the framework of the proposed Dual Contrastive Network (DCN), as shown in Fig. \ref{fig2}. It employs a Context-guided Contrastive Learning (CCL) and a Detail-guided Contrastive Learning (DCL) to explore the global contexts and local details, and the two branches achieve the mutual complementarity, thus enabling the model to learn more discriminative and detail information for different scene classification tasks.

\subsection{Problem Definition}
The FS-RSISC task follows the standard few-shot learning paradigm \cite{31,3}. In this paradigm, given a dataset ${\mathcal{D}}_{base}$ with sufficient samples of base classes, the goal is to enable the model to quickly adapt to novel tasks with few samples from the dataset ${\mathcal{D}}_{novel}$. Note that ${\mathcal{D}}_{base}$ and ${\mathcal{D}}_{novel}$ have no overlapped classes. 

Our model follows the transfer learning pipeline, which consists of a pre-training stage and a meta-test stage. In the pre-training stage, supervised learning is applied to train a feature encoder $f_{\theta}$ with ${\mathcal{D}}_{base}$. In the meta-test stage, the feature encoder $f_{\theta}$ is then transferred to the novel tasks for a better representation of ${\mathcal{D}}_{novel}$. In each novel task, there are a support set ${\mathcal{D}}_{S}$ and a query set ${\mathcal{D}}_{Q}$. The support set is formed by C classes selected from the ${\mathcal{D}}_{novel}$ with K samples per class, which is denoted as ${\mathcal{D}}_{S} = {\left\{(x_{i}, y_{i})\right\}}_{i=1}^{CK}$, where $(x_{i}, y_{i})$ is a pair of image and its ground-truth label. Likewise, the query set ${\mathcal{D}}_{Q} = {\left\{(x_{j}, y_{j})\right\}}_{j=1}^{CQ}$ is formed by C classes with Q samples per class. Based on this setting, the probability of a query sample $x \in {\mathcal{D}}_{Q}$ being predicted as class c is:
\begin{equation}\label{1}
{P}\left(y=c|x\right) = {softmax}\left(\langle f_{\theta} (x), w_{c} \rangle \right),
\end{equation}
where $\langle\cdot, \cdot\rangle$ is the distance function, $w_{c}={\frac{1}{K}}\sum_{x\in{\mathcal{D}_{S}^c}}f_{\theta}(x)$ is the prototype of the class c and $\mathcal{D}_{S}^c$ denotes the set of samples labeled with class c.

\begin{figure}[t]
\centering
\includegraphics[scale=0.5]{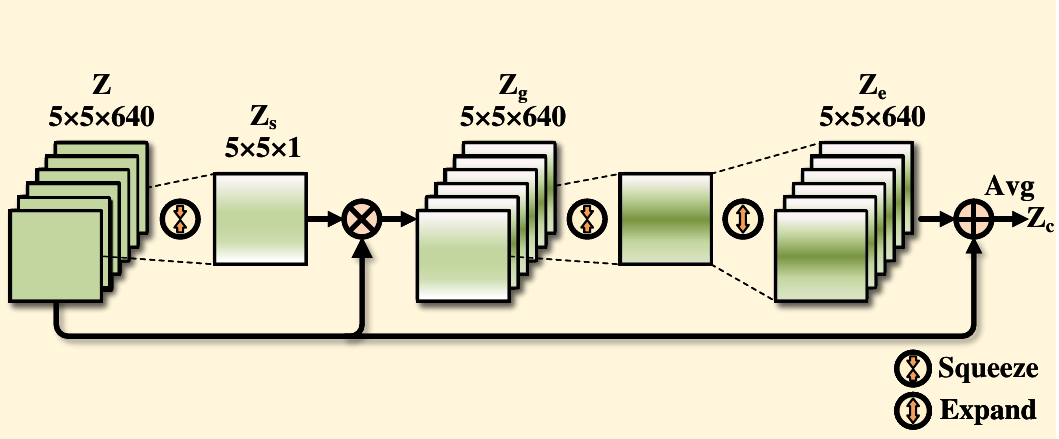}
\caption{Illustration of Condenser Network. With the Squeeze and Expand operations among channels, the scene-relevant information is enhanced while the irrelevant information is ignored. After the global average pooling, the context features are generated. }\label{fig3}
\end{figure}

\subsection{Context-guided Contrastive Learning}
One main challenge of FS-RSISC is the small inter-class variances. To address this challenge, we propose a Context-guided Contrastive Learning (CCL) to exploit the global contexts, which consists of a Condenser Network and a contrastive learning module. The goal is to enable the model to explore the discriminative features that can better reflect inter-class differences and thus guide the classification of similar images.

\subsubsection{Condenser Network}
Due to the interference of the background and irrelevant objects, some information relevant to the scene label in remote sensing images is always inapparent. To reduce this adverse effect, we design a Condenser Network for CCL to capture the global context information, as shown in Fig. \ref{fig3}. The specific process is as follows:

First, the feature maps $Z\in R^{h\times w\times c}$ output by the Feature Encoder are squeezed along the channel dimension using a 1 $\times$ 1 convolution layer to aggregate the spatial information from each channel. $Z_{s}\in R^{h\times w\times 1}$ is the obtained spatial map:
\begin{equation}\label{1}
\mathop{Z_{s}}=\mathop{f_{sq}(Z)}\hspace{1mm},
\end{equation}
where $f_{sq}\left(\cdot\right)$ is the Squeeze operation.

Then, each pixel point in $Z_{s}$ is mapped to 0-1 with the Softmax function and multiplied by the feature maps $Z$ in order to assign different weights to different spatial locations. The generated feature maps $Z_{g}\in R^{h\times w\times c}$ are formulated as:
\begin{equation}\label{1}
\mathop{Z_{g}}={\mathop{softmax}\left(Z_{s}\right)}{\times}{Z}.
\end{equation}

Next, the generated feature maps ${Z_{g}}$ are squeezed again to aggregate the features at each spatial position. The Layer Normalization and ReLU function are then employed on the squeezed feature maps to find out the vital regions while filtering out the insignificant ones. Subsequently, the squeezed feature maps are expanded by a 1 $\times$ 1 convolution layer with c kernels to convey vital spatial information to each channel, resulting in the expanded feature maps ${Z_{e}}\in R^{h\times w\times c}$. Finally, ${Z_{e}}$ are added to the original feature maps $Z$ to enhance the features in vital spatial locations. The enhanced feature maps $\tilde{Z}\in R^{h\times w\times c}$ are expressed as:
\begin{equation}\label{1}
\mathop{\tilde{Z}}=\mathop{f_{ex}}\left({ReLU}\left({LN}\left({f_{sq}\left(Z_{g}\right)}\right)\right)\right) + Z\hspace{1mm},
\end{equation}
where $f_{ex}\left(\cdot\right)$ is the Expand operation, $LN$ denotes the Layer Normalization.

After the global average pooling operation $avg$, the context features ${Z_{c}}\in R^{1\times 1\times c} = avg{(\tilde{Z})}\hspace{1mm}$ are obtained.

\subsubsection{Contrastive Learning}
In a training batch, we first implement two different data augmentations \cite{22} to achieve double-batch expansion. These two batch expansions are both propagated through the Feature Encoder and the Condenser Network to get the context features. Then, a non-linear projection network $f_{pc}\left(\cdot\right)$ and an L2 normalization layer are employed to map the context features to another embedding space, in which the contrastive learning is performed:
\begin{equation}\label{1}
F = f_{pc}(Z_{c})\hspace{1mm}, 
\end{equation}
where ${Z_{c}}\in R^{1\times 1\times c}$ are the context features.

Suppose there are $N$ samples in a batch, thus there are 2$N$ samples in the expanded batches. We set the sample $\left(x_i, y_i\right)$ as the anchor, and the positive and the negative samples are those with the same or different labels as the anchor, respectively. The Context-guided Contrastive Learning (CCL) loss is formulated as: 
\begin{equation}\label{1}
\mathop{L_{CCL}}=\mathop{-}{\sum_{i=1}^{2N}}{\frac{1}{P_i}}\!{\sum_{j=1,j\ne{i},y_i=y_j}^{2N}}  \!\!\!\!\!\! log{\frac{exp\left(d(F_i,F_j){/}\tau\right)}{\sum_{n=1,n\ne{i}}^{2N}{exp}\left(d(F_i,F_n){/}\tau\right)}}\hspace{1mm},
\end{equation}
where $P_{i}$ is the number of positive samples, $\tau$ is a scalar temperature parameter, $d(\cdot, \cdot)$ denotes the distance function, which is specifically implemented as a normalized dot product.

\subsection{Detail-guided Contrastive Learning}
Scene images in the same category sometimes are quite different, which results in a large intra-class variance. To alleviate this problem, we further propose a Detail-guided Contrastive Learning (DCL) approach with more attention on invariant local details in intra-class samples, which contains a Smelter Network and a contrastive learning module. 

\subsubsection{Smelter Network}
\begin{figure}[t]
\centering
\includegraphics[scale=0.5]{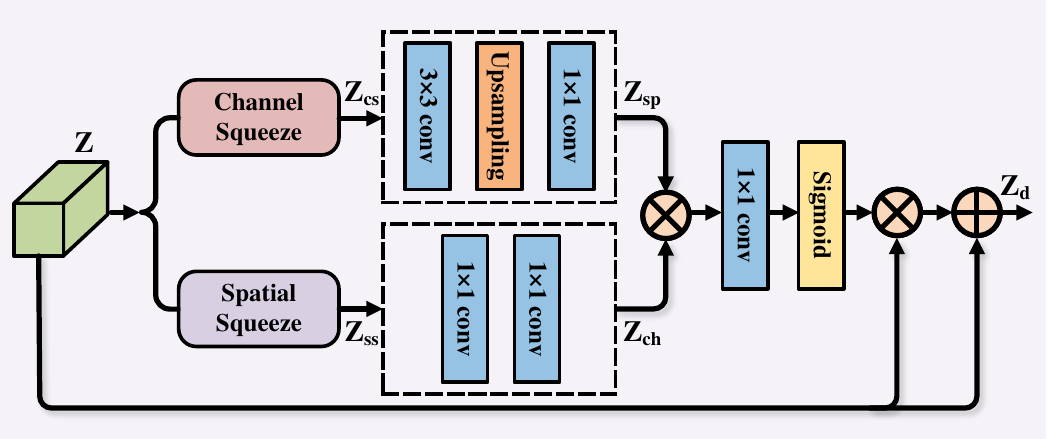}
\caption{Illustration of Smelter Network. The helpful information in the channel and spatial dimensions is extracted respectively to enhance the significant local detail regions of the feature maps.}\label{fig4}
\end{figure}

For DCL, we first propose a Smelter Network to locate and enhance the significant local detail information by self-adaptive adjustment, as illustrated in Fig.\ref{fig4}.

Specifically, the Channel Squeeze and Spatial Squeeze operations are first respectively performed on the feature maps $Z\in R^{h\times w\times c}$ by average pooling in the channel and spatial dimensions to obtain the channel squeezed map $Z_{cs}\in R^{h\times w\times 1}$ and the spatial squeezed map $Z_{ss}\in R^{1\times 1\times c}$:
\begin{equation}\label{1}
\mathop{Z_{cs}}=\mathop{f_{c}(Z)},                  
\end{equation}
\begin{equation}\label{1}
\mathop{Z_{ss}}=\mathop{f_{s}(Z)},
\end{equation}
where $f_{c}\left(\cdot\right)$ and $f_{s}\left(\cdot\right)$ are the Channel Squeeze operation and Spatial Squeeze operation.

The channel squeezed map $Z_{cs}$ is then successively input to a 3 $\times$ 3 convolution layer, an upsampling layer and a 1 $\times$ 1 convolution layer to derive the spatial attention map $Z_{sp}\in R^{h\times w\times 1}$: 
\begin{equation}\label{1}
\mathop{Z_{sp}}=\mathop{conv_{1\times 1}\left(up\left(conv_{3\times 3}(Z_{cs})\right)\right)}\hspace{1mm},  \end{equation}
where $conv_{n\times n}$ denotes an \emph{n} $\times$ \emph{n} convolution layer and $up$ denotes an upsampling layer.

Similarly, the spatial squeezed map $Z_{ss}$ is also propagated through two 1 $\times$ 1 convolution layers to get the channel attention map $Z_{ch}\in R^{1\times 1\times c}$:
\begin{equation}\label{1}
\mathop{Z_{ch}}=\mathop{conv_{1\times 1}\left(conv_{1\times 1}\left(Z_{ss}\right)\right)}.
\end{equation}

Next, the two attention maps are multiplied together and then mapped by a 1 $\times$ 1 convolution layer to obtain the synthetic attention maps. Finally, the synthetic attention maps are activated by the Sigmoid function and multiplied with the input feature maps $Z$, and then added to the feature maps $Z$, which achieves an enhancement of significant local detail regions. The obtained detail feature maps $Z_{d}\in R^{h\times w\times c}$ are formulated as:  
\begin{equation}\label{1}
\mathop{Z_{d}}=\mathop{\varepsilon}\left({conv_{1\times 1}}\left(Z_{sp}\times Z_{ch}\right)\right){\times}{Z}+Z \hspace{1mm},
\end{equation}
where $conv$ is a $1\times1$ convolution layer, $\varepsilon\left(\cdot\right)$ is the Sigmoid function.

\subsubsection{Contrastive Learning}
In a training batch, we implement the same data augmentations as CCL. The obtained two batch expansions are then propagated through the Feature Encoder and the Smelter Network to get the detail feature maps $Z_{d}\in R^{h\times w\times c}$. To better reflect the correlation among detail features and measure the distance, we further introduce a self-attention based feature alignment mechanism. The detail feature maps $Z_{d}$ are first mapped into three vectors by the projection network $f_{pf}\left(\cdot\right)$ to get the query $Q$, key $K$ and value $V$:
\begin{equation}\label{1}
Q = K = V = f_{pf}\left({Z}_{d}\right),
\end{equation}
where $Q$, $K$, $V$ $\in R^{hw\times \bar c}$.
Suppose given a pair of samples $\left(x_i, y_i\right)$ and $\left(x_j, y_j\right)$, the value $V_{i}$ is aligned with $V_{j}$ using the query $Q_{j}$ and the key $K_{i}$:
\begin{equation}\label{1}
V_{i\mid j} = \mathop{softmax}\left(\frac {{Q_j}{K_i^{\top}}}{\surd {\bar c}}\right){\times}{V_i}.
\end{equation}
Likewise, the value $V_{j}$ is aligned with $V_{i}$:
\begin{equation}\label{1}
V_{j\mid i} = \mathop{softmax}\left(\frac {{Q_i}{K_j^{\top}}}{\surd {\bar c}}\right){\times}{V_j}. 
\end{equation}
Therefore the distance between $x_i$ and $x_j$ is expressed as:
\begin{equation}\label{1}
\mathop{d(V_i,V_j)}=\mathop{\frac{1}{2}}\left(cos\left(V_{i}, V_{j\mid i}\right)+cos\left(V_{j}, V_{i\mid j}\right)\right),
\end{equation}

Similar to $L_{CCL}$, the Detail-guided Contrastive Learning (DCL) loss is formulated as:
\begin{equation}\label{1}
\mathop{L_{DCL}}=\mathop{-}{\sum_{i=1}^{2N}}{\frac{1}{P_i}}\!{\sum_{j=1,j\ne{i},y_i=y_j}^{2N}}  \!\!\!\!\!\! log{\frac{exp\left(d(V_i,V_j){/}\bar\tau\right)}{\sum_{n=1,n\ne{i}}^{2N}{exp}\left(d(V_i,V_n){/}\bar\tau\right)}}\hspace{1mm},
\end{equation}

Therefore, the total contrastive learning loss is composed of $L_{CCL}$ $\left(Eq.\left(6\right)\right)$ and $L_{DCL}$ $\left(Eq.\left(16\right)\right)$:
\begin{equation}\label{1}
\mathop{L_{CL}}=\mathop{L_{CCL}+\alpha L_{DCL}},
\end{equation}
where $\alpha$ is a hype-parameter to balance both items.

\subsection{Total Loss}
The cross-entropy loss is applied to optimize the distribution of both context and detail features spaces:
\begin{equation}\label{1}
\begin{aligned}
\mathop{\mathcal{L}_{CE}}=&\mathop{-}{\sum_{i=1}^{N}}{\frac{1}{N}}{\sum_{c=1}^{C}}{y_{ic}}log{\frac{exp({o}_{ic})}{\sum_{j=1}^{C}{exp}({o}_{jc})}}\\&{-}{\beta}{\sum_{i=1}^{N}}{\frac{1}{N}}{\sum_{c=1}^{C}}{y_{ic}}log{\frac{exp(\tilde{o}_{ic})}{\sum_{j=1}^{C}{exp}(\tilde{o}_{jc})}}\hspace{1mm},
\end{aligned}
\end{equation}
where $N$ is the number of samples in a training batch, ${y_{ic}}$ equals to 1 when the sample i belongs to class c, $C$ is the number of categories, $\beta$ is a hype-parameter to balance the two items, and $o$ and $\tilde{o}$ represent the logits output by the last layer of the Condenser Network and the Smelter Network, respectively.

Overall, our full pre-training objective is to minimize the total loss, which consists of the cross-entropy loss $L_{CE}$ $\left(Eq.\left(18\right)\right)$ and the contrastive learning loss $L_{CL}$ $\left(Eq.\left(17\right)\right)$:
\begin{equation}\label{1}
\mathop{L_{total}}=\mathop{L_{CE}+\gamma L_{CL}}\hspace{1mm},
\end{equation}
where $\gamma$ is a hype-parameter to balance each item.

\subsection{Test}
In each task of the meta-test stage, a support set 
${\mathcal{D}}_{S} = {\left\{(x_{i}, y_{i})\right\}}_{i=1}^{CK}$ is sampled from $\mathcal{D}_{novel}$. Then, there are two choices of prototypes of class $c$:
\begin{equation}
    w_c = \frac{1}{K} \sum_{i=1}^{K} {Z}_i^c, and \hspace{1mm} \tilde{w}_c = \frac{1}{K} \sum_{i=1}^{K} avg(\tilde{Z}_i^c),
\end{equation}
where $Z_i^c$ and $\tilde{Z}_i^c$ represent the context features $Z_c$ and the detail feture maps $Z_d$ of the $i$-th sample in the $c$-th class, respectively.

$\mathcal{D}_Q$ denotes a query set sampled from $\mathcal{D}_{novel}$. Then, the probability of a query sample $x_q\in \mathcal{D}_Q$ being predicted as class $c$ is:

\begin{small}
\begin{equation}
{P}\!\left(y\!\!=\!\!c|x_q\right) \!\!=\!\!\frac{1}{2}({softmax}(\langle Z_q, w_{c} \rangle)+{softmax}(\langle avg(\tilde{Z}_q), \tilde{w}_{c} \rangle )).
\end{equation}
\end{small}

Besides, note that the Feature Encoder, the Condenser Network and the Smelter Network are frozen, and the two contrastive learning modules are discarded in the meta-test stage.

\section{Experiments}
\subsection{Datasets}
We evaluate the proposed DCN on four popular remote sensing datasets: WHU-RS19 \cite{25}, UC Merced \cite{26}, NWPU-RESISC45 \cite{24} and AID \cite{27}. WHU-RS19 and UC Merced are small datasets, while NWPU-RESISC45 and AID are large datasets. According to the division criteria of few-shot learning \cite{28}, the datasets are divided into training set, validation set and test set with disjoint classes. Details of these datasets are described below.

WHU-RS19 dataset \cite{25} is collected by Whuhan University from Google Earth. It contains 1,005 images of 600 $\times$ 600 in 19 classes. And each class includes at least 50 images. Following \cite{10}, WHU-RS19 is divided into 9, 5, and 5 classes for training, validation, and testing, respectively.

UC Merced dataset \cite{26} is released by University of California, Merced. It includes 100 images of 21 classes, pixel size of per image is 256 $\times$ 256. Following \cite{10}, UC Merced is split into 10, 6, and 5 classes for training, validation, and testing, respectively.

NWPU-RESISC45 dataset \cite{24} is a large dataset released by Northwestern Polytechnical University for RSISC, which contains 31,500 images with 256 $\times$ 256 pixels. There are 45 scene classes, and each class has 700 images. Following \cite{10}, NWPU-RESISC45 is divided into a training set with 25 classes, a validation set with 10 classes and a test set with 10 classes.

AID dataset \cite{27} contains 10,000 images of 30 classes, which issued by Wuhan University and Huazhong University of Science and Technology. Each class has 220 $ \sim $ 420 samples and each image is 600 $\times$ 600 pixels. Following \cite{10}, AID is split into a training set with 16 classes, a validation set with 7 classes and a test set with 7 classes.

\begin{table*}[t]
\centering
\caption {Comparative results(\%) on the WHU-RS19 and UC Merced datasets with $\pm$95$\%$ confidence intervals. The top two results are shown in bold and underline. }
\setlength{\tabcolsep}{3mm}{
\begin{tabular}{ccccccc}
\toprule
\multirow{2}{*} {Type} & \multirow{2}{*}{Method} & \multirow{2}{*}{Backbone} & \multicolumn{2}{c}{ WHU-RS19} & \multicolumn{2}{c}{UC Merced} \\ 
\cmidrule(r){4-5}  \cmidrule(r){6-7}  
~ & ~ & ~ & 1-shot & 5-shot & 1-shot & 5-shot \\ \midrule
\multirow{2}{*}{Optimization} & LLSR\cite{55} & Conv & 57.10 & 70.65 & 39.47 & 57.40 \\
 & MAML\cite{17} & ResNet-12 & 59.19$\pm$0.92 & 72.34$\pm$0.75 & 50.23$\pm$0.98 & 62.02$\pm$0.72 \\ \hline
 \multirow{8}{*}{Metric} & DLA-MatchNet\cite{52} & Conv & 68.27$\pm$1.83 & 79.89$\pm$0.33 & 53.76$\pm$0.62 & 63.01$\pm$0.51 \\
 & SCL-MLNet\cite{53} & Conv & {-} & {-} & 51.37$\pm$0.79 & 68.09$\pm$0.92 \\
 & MatchingNet\cite{69} & ResNet-12 & 75.99$\pm$0.63 & 82.96$\pm$0.40 & 50.99$\pm$0.69 & 57.14$\pm$0.59 \\ 
 & ProtoNet\cite{19} & ResNet-12 & 65.53$\pm$0.99 & 85.53$\pm$0.73 & 53.15$\pm$0.80 & 71.54$\pm$0.52 \\
 & RelationNet\cite{20} & ResNet-12 & 65.36$\pm$0.75 & 80.02$\pm$0.64 & 49.18$\pm$0.78 & 63.64$\pm$0.54 \\
 & deepEMD\cite{66} & ResNet-12 & 67.37$\pm$0.62 & 85.43$\pm$0.43 & 52.11$\pm$0.41 & 69.82$\pm$0.36 \\
 & FRN\cite{67} & ResNet-12 & 63.75$\pm$0.19 & 83.18$\pm$0.12 & 51.14$\pm$0.22 & 65.46$\pm$0.15 \\
 & PSM\cite{83} & ResNet-12 & 67.72$\pm$0.95 & 86.21$\pm$0.70 & 54.35$\pm$0.82 & 71.86$\pm$0.54 \\ 
 & IDLN\cite{9} & ResNet-12 & 73.89$\pm$0.88 & 83.12$\pm$0.56 & {-} & {-} \\
 & SPNet\cite{10} & ResNet-18 & \underline{81.06$\pm$0.60} & \underline{88.04$\pm$0.28} & \underline{57.64$\pm$0.73} & \underline{73.52$\pm$0.51} \\
 \hline
 Transfer & S2M2\cite{11} & ResNet-12 & 69.00$\pm$0.41 & 82.14$\pm$0.21 & 52.16$\pm$0.46 & 67.38$\pm$0.25 \\
 & DeepDBC\cite{12} & ResNet-12 & 68.23$\pm$0.39 & 85.90$\pm$0.31 & 54.20$\pm$0.35 & 71.94$\pm$0.26 \\
 \hline
Transfer & \textbf{DCN (Ours)} & ResNet-12 & \textbf{81.74$\pm$0.55} & \textbf{91.67$\pm$0.25} & \textbf{58.64$\pm$0.71} & \textbf{76.61$\pm$0.49}\\
\bottomrule
\end{tabular}}
\label{table1}
\end{table*}

\begin{table*}[t]
\centering
\caption {Comparative results(\%) on the NWPU-RESISC45 and AID datasets with $\pm$95$\%$ confidence intervals. The top two results are shown in bold and underline.}
\setlength{\tabcolsep}{3mm}{
\begin{tabular}{ccccccc}
\toprule
\multirow{2}{*}{Type} & \multirow{2}{*}{Method} & \multirow{2}{*}{Backbone} & \multicolumn{2}{c}{ NWPU-RESISC45} & \multicolumn{2}{c}{AID} \\ 
\cmidrule(r){4-5}  \cmidrule(r){6-7} 
~ & ~ & ~ & 1-shot & 5-shot & 1-shot & 5-shot \\
\midrule
\multirow{2}{*}{Optimization} & LLSR\cite{55} & Conv & 51.43 & 72.90 & 45.18 & 61.76 \\
 & MAML\cite{17} & ResNet-12 & 56.01$\pm$0.87 & 72.94$\pm$0.63 & 62.54$\pm$0.79 & 73.12$\pm$0.85 \\ \hline
 \multirow{8}{*}{Metric} & DLA-MatchNet\cite{52} & Conv & 68.80$\pm$0.70 & 81.63$\pm$0.46 & 57.21$\pm$0.82 & 73.45$\pm$0.61 \\
 & SCL-MLNet\cite{53} & Conv & 62.21$\pm$1.12 & 80.86$\pm$0.76 & 59.46$\pm$0.96 & 76.31$\pm$0.68 \\
 & MatchingNet\cite{69} & ResNet-12 & 61.57$\pm$0.49 & 76.02$\pm$0.34 & 64.30$\pm$0.46 & 74.49$\pm$0.35 \\ 
 & ProtoNet\cite{19} & ResNet-12 & 64.52$\pm$0.48 & 81.95$\pm$0.30 & 67.08$\pm$0.47 & 82.44$\pm$0.29 \\
 & RelationNet\cite{20} & ResNet-12 & 65.52$\pm$0.85 & 78.38$\pm$0.31 & 68.56$\pm$0.49 & 79.21$\pm$0.35 \\
 & deepEMD\cite{66} & ResNet-12 & 67.50$\pm$0.74 & 80.54$\pm$0.51 & 63.78$\pm$0.68 & 80.26$\pm$0.41 \\
 & FRN\cite{67} & ResNet-12 & 66.31$\pm$0.20 & 79.67$\pm$0.15 & 66.97$\pm$0.23 & 78.82$\pm$0.16 \\
 & PSM\cite{83} & ResNet-12 & 66.17$\pm$0.51 & 83.03$\pm$0.34 & \underline{68.89$\pm$0.46} & 83.75$\pm$0.33 \\ 
 & IDLN\cite{9} & ResNet-12 & \textbf{75.25$\pm$0.75} & 84.67$\pm$0.23 & {-} & {-} \\
 & SPNet\cite{10} & ResNet-18 & 67.84$\pm$0.87 & 83.94$\pm$0.50 & {-} & {-} \\
 \hline
 \multirow{2}{*}{Transfer} & S2M2\cite{11} & ResNet-12 & 63.24$\pm$0.47 & 83.23$\pm$0.28 & 66.22$\pm$0.45 & 82.87$\pm$0.29 \\
 & DeepDBC\cite{12} & ResNet-12 & 68.84$\pm$0.44 & 84.12$\pm$0.29 & 66.56$\pm$0.43 & \underline{84.15$\pm$0.25} \\
 & CSDL\cite{41} & ResNet-12 & 71.28$\pm$0.72 & \underline{85.66$\pm$0.47} & {-} & {-} \\
 \hline
Transfer & \textbf{DCN (Ours)} & ResNet-12 & \underline{74.40$\pm$0.78} & \textbf{89.22$\pm$0.41} & \textbf{73.38$\pm$0.77} & \textbf{88.25$\pm$0.44} \\
\bottomrule
\end{tabular}}
\label{table1}
\end{table*}

\subsection{Experiment Setup}
Following previous work \cite{30}, ResNet-12 is employed as the backbone for all the four benchmark datasets. For the projection network, $f_{pg}\left(\cdot\right)$ is a Multilayer Perceptron (MLP) with a hidden layer of size 640 and a ReLU layer, which outputs 128-dimensional vectors, and $f_{pl}\left(\cdot\right)$ is an MLP with a single layer of size 128 and a ReLU layer. In the data preprocessing stage, the images are all cropped to a fixed size 84 $\times$ 84. During the model pre-training stage, we apply Momentum-SGD for optimization, and set the learning rate as 0.05, the momentum as 0.9, and the weight decay as 0.0005. In terms of the loss weights, we set $\alpha$=$\gamma$=1, $\beta$=0.1. The classification accuracies are calculated by random selecting 600 tasks from the test set with a confidence interval of 95{\%}. Our model is implemented with the PyTorch framework and trained on a NVIDIA GeForce RTX 2080 GPU.

\subsection{Comparison with State-of-the-Art Methods}
We choose nine typical few-shot learning methods and six state-of-the-art FS-RSISC methods for comparison. The typical few-shot learning methods include one optimization-based approach \cite{5}, six metric-based approaches \cite{6,7,8,79,80,81}, and two transfer learning-based approaches \cite{29,82}. For a fair comparison, all methods utilize the ResNet-12 as the backbone. The results of MAML \cite{5}, Relation Networks \cite{8} are borrowed from the published articles \cite{2,12} and the others are re-implemented by us according to \cite{3,29}. The state-of-the-art FS-RSISC methods can be divided into optimization-based approach LLSR \cite{9}, metric-based approaches, such as DLA-MatchNet \cite{10}, SCL-MLNet \cite{11}, IDLN \cite{12}, and SPNet \cite{1}, and transfer-based approach CSDL \cite{2}. All results are cited from the original published articles. 

\begin{table}[t]
\centering
\caption {Ablation studies(\%) on the WHU-RS19 and UC Merced datasets.}
\setlength{\tabcolsep}{3mm}{
\scalebox{0.8}{
\begin{tabular}{ccccc}
\toprule
\multirow{2}{*}{Method} & \multicolumn{2}{c}{WHU-RS19} & \multicolumn{2}{c}{UC Merced} \\
\cmidrule(r){2-3}  \cmidrule(r){4-5} 
~ & 1-shot & 5-shot & 1-shot & 5-shot\\ \midrule
Backbone & 67.65$\pm$0.74 & 83.17$\pm$0.36 & 51.87$\pm$0.70 & 68.90$\pm$0.55 \\
CCL (w/o CL) & 68.17$\pm$0.72 & 84.08$\pm$0.38 & 52.52$\pm$0.72 & 70.02$\pm$0.52 \\
CCL (w/o CN) & 67.13$\pm$0.66 & 83.80$\pm$0.37 & 54.37$\pm$0.75 & 72.40$\pm$0.48 \\
CCL & 69.03$\pm$0.67 & 85.22$\pm$0.40 & 54.93$\pm$0.68 & 72.86$\pm$0.50 \\
DCL (w/o CL) & 67.61$\pm$0.69 & 83.11$\pm$0.36 & 52.10$\pm$0.65 & 69.57$\pm$0.58 \\
DCL (w/o SN) & 74.29$\pm$0.70 & 89.16$\pm$0.43 & 55.49$\pm$0.66 & 73.33$\pm$0.45 \\ 
DCL & 75.58$\pm$0.62 & 89.75$\pm$0.35 & 56.52$\pm$0.72 & 74.29$\pm$0.52\\
\textbf{DCN (Ours)} & \textbf{81.74$\pm$0.55} & \textbf{91.67$\pm$0.25} & \textbf{58.64$\pm$0.71} & \textbf{76.61$\pm$0.49}\\
\bottomrule
\end{tabular}}}
\label{table1}
\end{table}

\begin{table}[t]
\centering
\caption {Ablation studies(\%) on the NWPU-RESISC45 and AID datasets.}
\setlength{\tabcolsep}{3mm}{
\scalebox{0.8}{
\begin{tabular}{ccccc}
\toprule
\multirow{2}{*}{Method} & \multicolumn{2}{c}{NWPU-RESISC45} & \multicolumn{2}{c}{AID} \\
\cmidrule(r){2-3}  \cmidrule(r){4-5} 
~ & 1-shot & 5-shot & 1-shot & 5-shot\\ \midrule
Backbone & 71.26$\pm$0.86 & 86.19$\pm$0.47 & 69.22$\pm$0.79 & 83.48$\pm$0.45 \\
CCL (w/o CL) & 71.66$\pm$0.85 & 86.31$\pm$0.48 & 69.35$\pm$0.84 & 83.07$\pm$0.49 \\
CCL (w/o CN) & 72.01$\pm$0.82 & 87.20$\pm$0.44 & 69.18$\pm$0.79 & 84.46$\pm$0.45 \\
CCL & 72.55$\pm$0.71 & 87.92$\pm$0.41 & 70.91$\pm$0.80 & 85.66$\pm$0.44 \\
DCL (w/o CL) & 71.91$\pm$0.75 & 86.53$\pm$0.47 & 69.65$\pm$0.82 & 84.10$\pm$0.51 \\
DCL (w/o SN) & 72.59$\pm$0.71 & 87.47$\pm$0.45 & 71.12$\pm$0.79 & 86.27$\pm$0.46 \\ 
DCL & 73.50$\pm$0.76 & 88.37$\pm$0.47 & 72.16$\pm$0.77 & 86.97$\pm$0.48\\
\textbf{DCN (Ours)} & \textbf{74.40$\pm$0.78} & \textbf{89.22$\pm$0.41} & \textbf{73.38$\pm$0.77} & \textbf{88.25$\pm$0.44}\\
\bottomrule
\end{tabular}}}
\label{table1}
\end{table}

\subsubsection{Results on samll datasets}
Table I shows the comparative results on the small datasets: WHU-RS19 \cite{25} and UC Merced \cite{26}, where CSDL has no results on both datasets. We observe that our method achieves the best results on both 5-way 1-shot and 5-way 5-shot. Concretely, on 1-shot setting, our DCN achieves accuracies of 81.74{\%} on the WHU-RS19 dataset and 58.64{\%} on the UC Merced dataset, which are at least 0.68{\%} and 1{\%} higher than the other methods, respectively. On 5-shot setting, the DCN outperforms the sub-optimal approach SPNet by a large margin of 3.63{\%} and 3.09{\%} on the two datasets, respectively. This indicates that both context and detail features of remote sensing images contribute to scene classification, and our DCN is able to learn more robust and comprehensive features with the dual contrastive learning framework, which can better adapt to the different novel scene classification tasks. In addition, it is obvious that the gains achieved on 5-shot setting are greater than those on 1-shot setting. This is due to the fact that unlike meta-learning methods that enhance the learning ability of the model and enable the model learn how to learn, our DCN focuses more on feature representations to obtain more discriminative features.

\subsubsection{Results on large datasets}
On the large datasets, NWPU-RESISC45 \cite{24} and AID \cite{27}, our DCN also achieves competitive performance. From Table II, we can observe that our method achieves encouraging improvements over the state-of-the-art approaches on 5-way 5-shot, where the gains of 3.56{\%} and 4.1{\%} are achieved than the second best approach on the NWPU-RESISC45 and AID datasets, respectively. On 5-way 1-shot, our method is 0.85{\%} lower than the best result in IDLN \cite{12} on the NWPU-RESISC45 dataset. We suppose the reason lies in that our DCN trains the classifier based on the support set features, while IDLN calibrates features iteratively to guide the generation of the classifier. Therefore our method may has the classification bias when the number of support samples is too small. Moreover, we observe that SPNet, which performs well on the small datasets, suffers from performance degradation when faced with the large datasets. Our DCN maintains superior performance on both large and small datasets, which demonstrates the generality of our method.

\subsection{Ablation Studies}
To demonstrate the effectiveness of each component of our DCN, we conduct a series of ablation studies. Our baseline is based on \cite{14}, which only applies the cross-entropy loss to learn the backbone in the pre-training stage. Then we consider the following variants:

\textbf{Backbone} represents the baseline of our method.

\textbf{CCL (w/o CL)} adds the Condenser Network after the Feature Encoder on the baseline.

\textbf{CCL (w/o CN)} adds the Context-guided Contrastive
Learning without the Condenser Network on the baseline. 

\textbf{CCL} adds the Context-guided Contrastive
Learning on the baseline.

\textbf{DCL (w/o CL)} adds the Smelter Network after the Feature Encoder on the baseline.

\textbf{DCL (w/o SN)} adds the Detail-guided Contrastive Learning without the Smelter Network on the baseline.

\textbf{DCL} adds the Detail-guided Contrastive
Learning on the baseline. 

The detailed results of ablation studies on the four benchmark datasets are shown in Table III and Table IV. The results show a significant improvement of our DCN over the baseline, achieving 3.14{\%} $ \sim $ 14.09{\%} gains on 1-shot setting and 3.03{\%} $ \sim $ 8.5{\%} gains on 5-shot setting. The CCL consists of the Condenser Network and the corresponding contrastive learning module. We can observe that CCL improves the results by 0.54{\%} $ \sim $ 1.9{\%} and 0.46{\%} $ \sim $ 1.42{\%} compared with CCL (w/o CN) on the two settings, respectively. It is demonstrated that the Condenser Network can generate features containing more discriminative information by exploiting the global contexts, thus enabling the model to pay more attention to inter-class variability during the CCL. Likewise, the Detail-guided Contrastive Learning consists of the Smelter Network and the corresponding contrastive learning module. It is observed that DCL outperforms DCL (w/o SN). Especially on the NWPU-RESISC45 and UC Merced datasets, DCL achieves about 1{\%} improvements on both settings. This proves the effectiveness of our designed Smelter Network, which guides the network to focus more on significant detail regions in DCL. Compared with CCL and DCL, which are both a single contrastive learning framework, our DCN achieves at least 0.9{\%} and 0.85{\%} performance gain on 5-way 1-shot and 5-way 5-shot, respectively. This demonstrates that our dual contrastive learning framework is able to learn robust feature representations by the complementarity of context and detail branches.

\begin{figure}[t]
\centering
\includegraphics[scale=0.55]{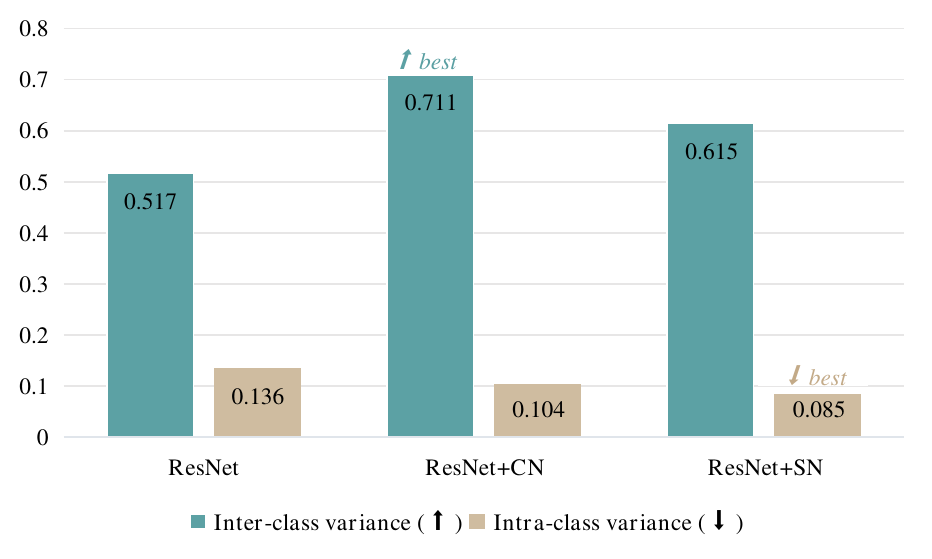}
\caption{The inter-class variances and intra-class variances on the NWPU-RESISC45 dataset on 5-way 1-shot with baseline (ResNet), Condenser Network-integrated (ResNet+CN) and Smelter Network-integrated (ResNet+SN).}\label{fig5}
\end{figure}

\renewcommand{\floatpagefraction}{.9}
\begin{figure*}[htbp]
\centering
\includegraphics[scale=0.41]{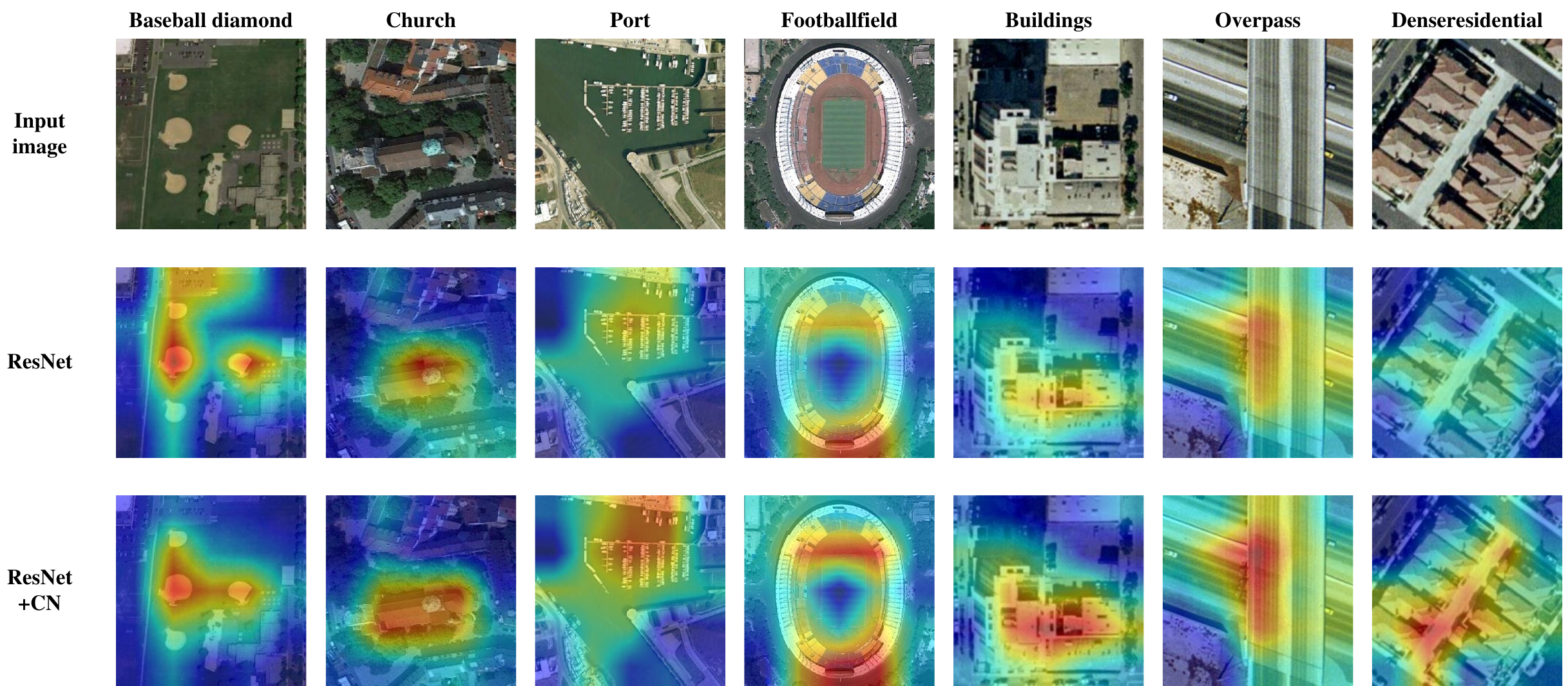}
\caption{Grad-CAM visualization results of inter-class images with different networks: baseline (ResNet) and Condenser Network-integrated (ResNet+CN).}\label{fig6}
\end{figure*}
\renewcommand{\floatpagefraction}{.9}
\begin{figure*}[htbp]
\centering
\includegraphics[scale=0.41]{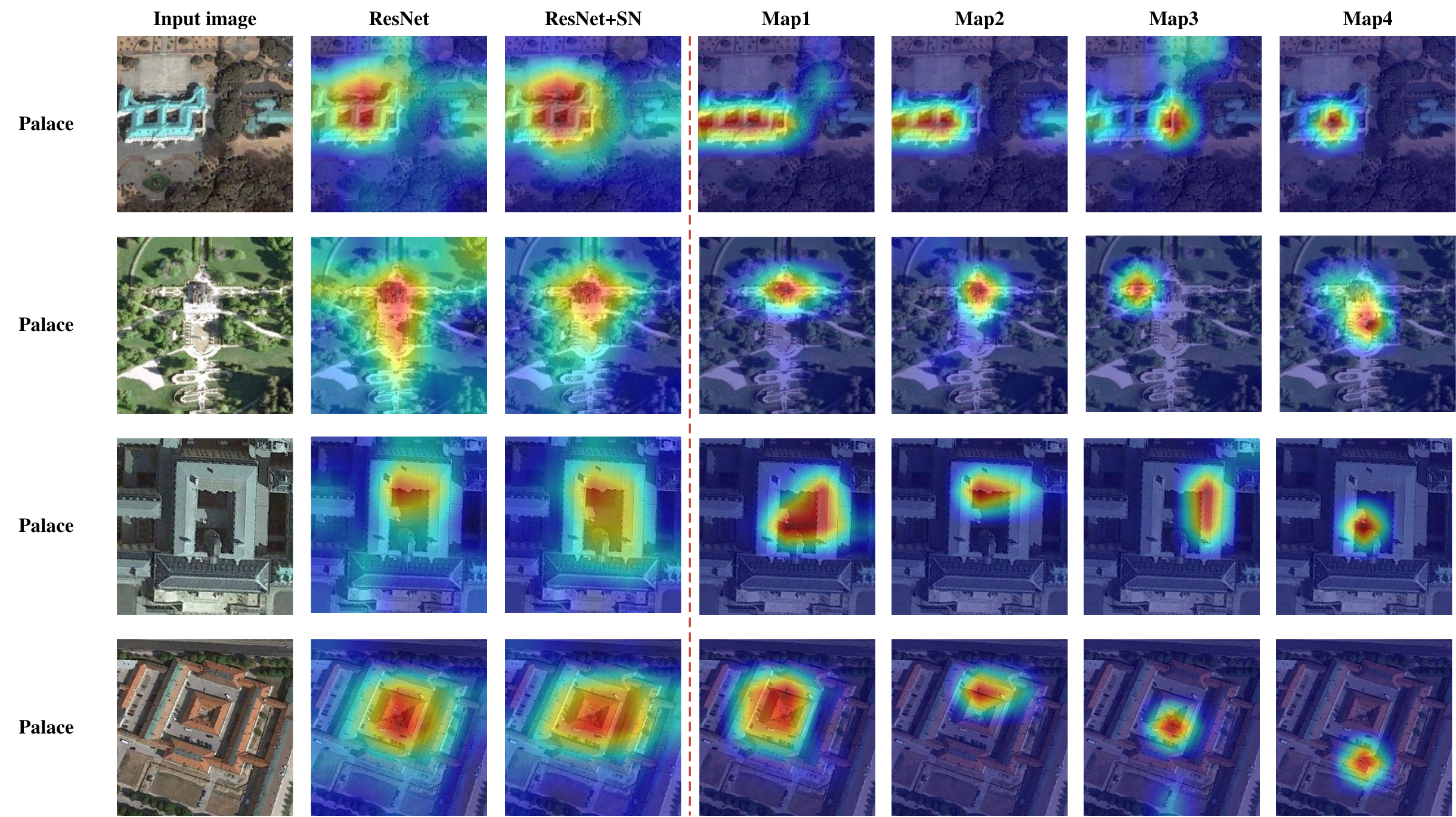}
\caption{Grad-CAM visualization results of intra-class images with baseline (ResNet) and Smelter Network-integrated (ResNet+SN), where the right of the
red dotted line is the visualization results of feature maps from several channels with large weights extracted by ResNet+SN.}\label{fig7}
\end{figure*}


\subsection{Inter-class and Intra-class Variances Analysis}
To verify the effectiveness of our proposed CCL and DCL, we visualize the inter-class and intra-class variances of novel classes on the NWPU-RESISC45 dataset with different networks, including baseline (ResNet) and two branches of our DCN, namely Condenser Network-integrated (ResNet+CN) and Smelter Network-integrated (ResNet+SN). As shown in Fig. \ref{fig5}, inter-class variance of embedding features by ResNet+CN is larger than that by ResNet, which indicates that the Condenser Network and auxiliary contrastive learning task based on the context features promote the model to extract more representative and discriminative features. This enlarge the distance between samples of different categories in the feature space to some extent. It is also observed that compared with ResNet, the features extracted by ResNet+SN has a smaller intra-class variance. This demonstrates that the Smelter Network and auxiliary contrastive learning task based on the detail features enable the model to focus more on the local details, which alleviates the problem of widely various appearances and backgrounds of the same class samples.



Moreover, as shown in Fig. \ref{fig6} and Fig. \ref{fig7}, Gradient-weighted Class Activation Mapping (Grad-CAM) \cite{73} is applied to visualize the interested regions of different networks for inter-class and intra-class images, where the regions of interest are indicated in red. From Fig. \ref{fig6}, it is observed that the representative regions, which are directly related to the scene categories, are more prominently highlighted in the third row. For example, for the baseball diamond in the first column, the Grad-CAM result of ResNet+CN highlights all four baseball diamonds, while the result of ResNet highlights some unrelated regions such as the limestone on the upper and right sides of the image. This indicates that the Condenser Network enhances the representation of relevant features while ignoring the redundant features, and promotes the model to learn more discriminative features with the contrastive learning. Comparing the second and third columns in Fig. \ref{fig7}, we observe that the regions concerned by the network are more comprehensive and accurate in the third column. This is due to the fact that DCL enables the network to pay attention to different spatial details on each channel. Therefore, the focal regions are magnified when all channels are aggregated together. To better demonstrate the attention of ResNet+SN to local detail regions, we additionally selected a few feature maps from channels that account for large weights. It is observed that each feature map extracted by ResNet+SN concentrates on important local information, thus the total class activation map contains more local details, which proves that the DCL complements the CCL for a better classification.


\subsection{Weighting Coefficients Analysis}

\begin{figure}[t]
	\centering
	\subfloat[Results on 5-way 1-shot.]
	{\includegraphics[scale=0.278]{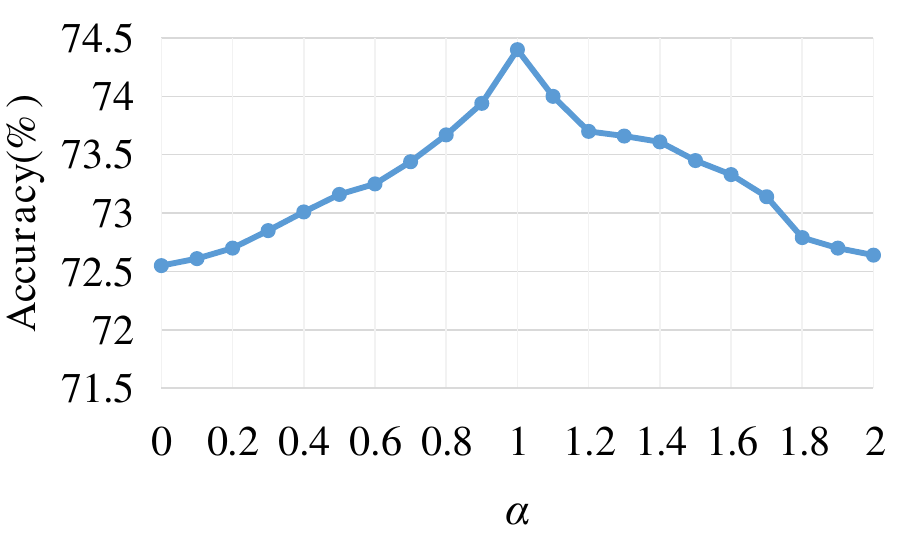}}\quad
    \subfloat[Results on 5-way 5-shot.]
	{\includegraphics[scale=0.278]{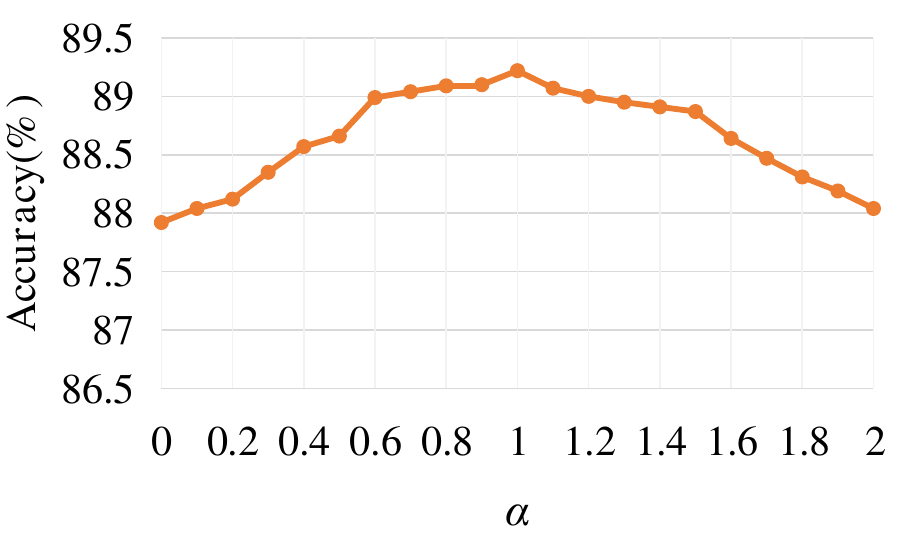}}
	\caption{Impact of weighting coefficient $\alpha$ on the NWPU-RESISC45 dataset.}
	\label{fig8}
\end{figure}

\begin{figure}[t]
	\centering
	\subfloat[Results on 5-way 1-shot.]
	{\includegraphics[scale=0.278]{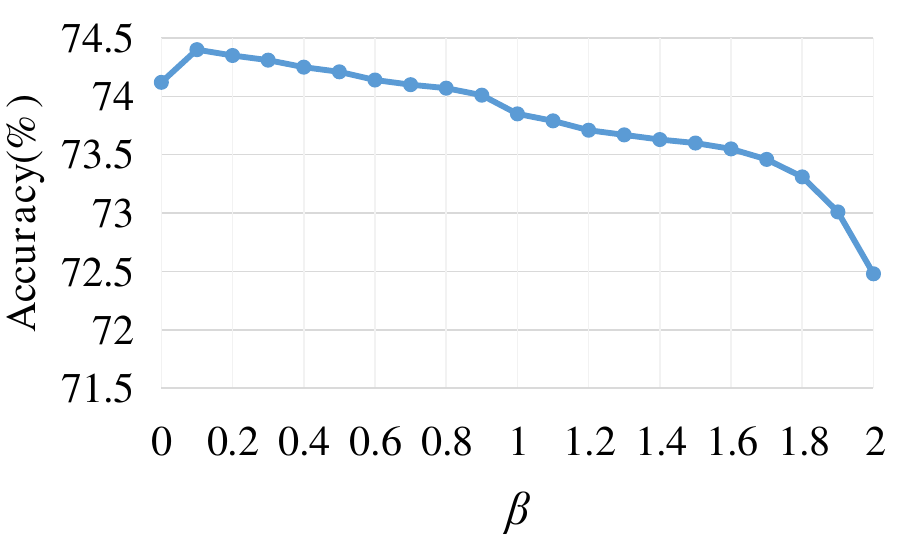}}\quad
    \subfloat[Results on 5-way 5-shot.]
	{\includegraphics[scale=0.278]{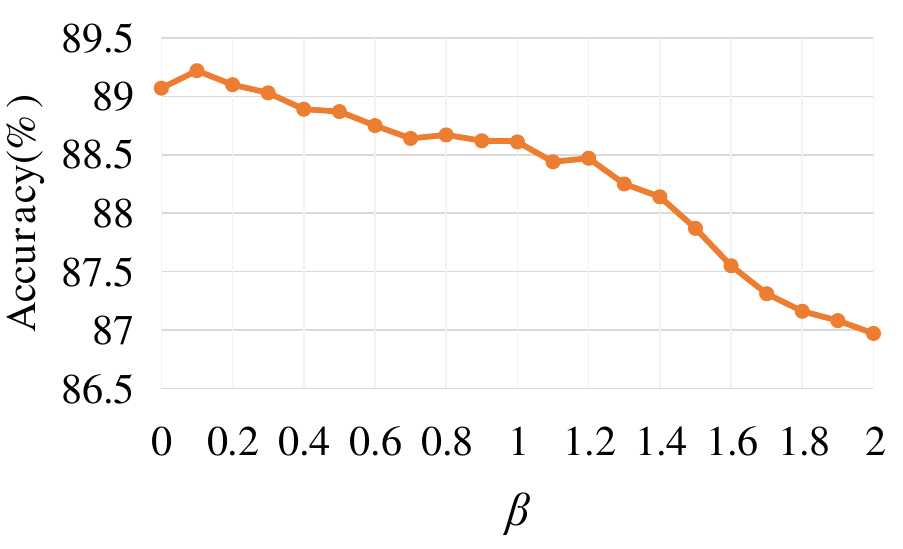}}
	\caption{Impact of weighting coefficient $\beta$ on the NWPU-RESISC45 dataset.}
	\label{fig9}
\end{figure}

\begin{figure}[t]
	\centering
	\subfloat[Results on 5-way 1-shot.]
	{\includegraphics[scale=0.278]{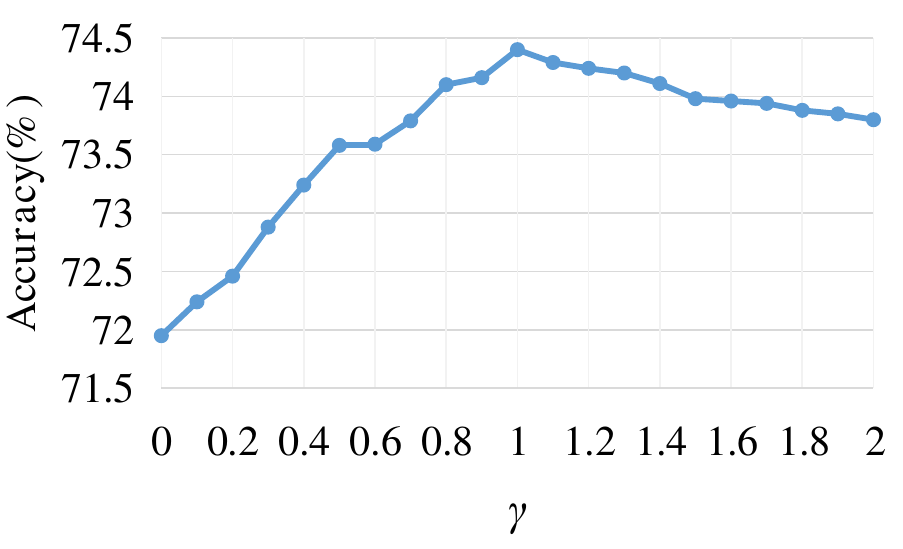}}\quad
    \subfloat[Results on 5-way 5-shot.]
	{\includegraphics[scale=0.278]{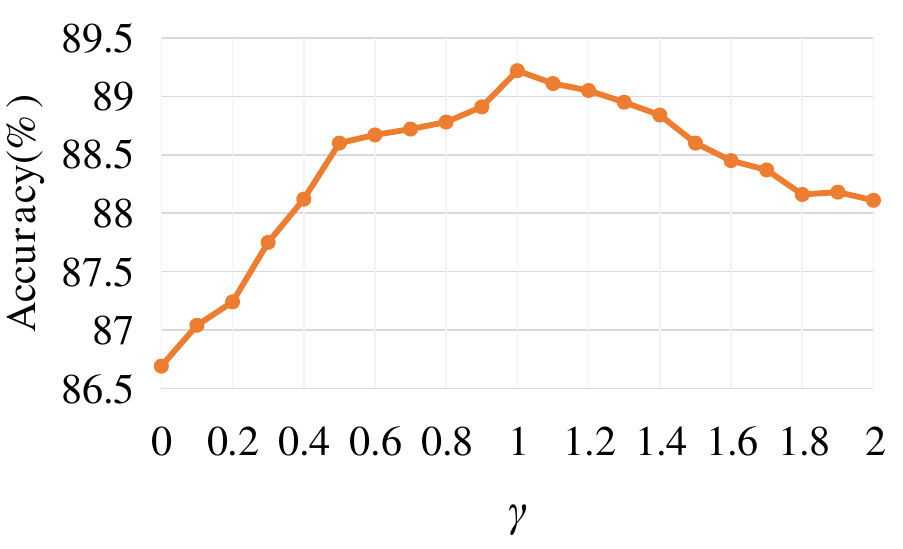}}
	\caption{Impact of weighting coefficient $\gamma$ on the NWPU-RESISC45 dataset.}
	\label{fig10}
\end{figure}

\subsubsection{Impact of parameter $\alpha$ sensitivity}
In this section, we investigate the impact of the weight coefficient $\alpha$ in Eq.(17) on the overall model performance, which balances the DCL and CCL in the dual contrastive learning task. In this process, we fix the weighting coefficients $\beta$ and $\gamma$ to 0.1 and 1, respectively. The results on the NWPU-RESISC45 dataset under different settings are shown in Fig. \ref{fig8}. It is observed that on both settings the performance reaches their peaks when $\alpha$ = 1 and decreases either $\alpha$ is greater or less than 1. This suggests that the DCL task and the CCL task are equally important in the dual contrastive learning, and their two losses complement each other to facilitate the learning of robust feature representations.

\subsubsection{Impact of parameter $\beta$ sensitivity}
The weight coefficient $\beta$ in Eq.(18) is applied to balance the two cross-entropy losses ratios of the context and detail features. We adjust $\alpha$ by fixing both the weighting coefficients $\alpha$ and $\gamma$ to 1. The results on the NWPU-RESISC45 dataset are shown in Fig. \ref{fig9}. As observed from the figure, when $\beta$ = 0.1, the performance of our DCN reaches the maximum and decreases as $\beta$ increasing over 0.1. And when no detail cross-entropy loss is added, i.e., $\beta$ = 0, there is also a small degradation in performance, which proves that the detail features are effective in classification, and can compensate for the bias introduced by training the classifier using only context features.

\subsubsection{Impact of parameter $\gamma$ sensitivity}
Our pre-training model is a multi-task learning network, which contains a classification task and a dual contrastive learning task. The weight coefficient $\gamma$ in Eq.(19) represents the proportion of the dual contrastive learning task in the multi-tasks. We fix $\alpha$ and $\beta$ to 1 and 0.1, respectively, and adjust $\gamma$. The results on the NWPU-RESISC45 dataset are shown in Fig. \ref{fig10}. We observe that the performance improves with the increase of $\gamma$ on both the 1-shot and 5-shot settings, and peak at $\gamma$ = 1, then continue to decrease. This indicates that dual contrastive learning has a positive impact on the overall model as an auxiliary task for the classification task, and both contrastive and classification tasks promote each other to learn a good feature embedding. 

\subsection{Training Time and Parameters}
\begin{table}[h]
\centering
\caption {Training runtime on the WHU-RS19, UC Merced, NWPU-RESISC45 and AID datasets.}
\setlength{\tabcolsep}{3mm}{
\scalebox{0.8}{
\begin{tabular}{ccccc}
\toprule
~ & WHU-RS19 & UC Merced & NWPU-RESISC45 & AID\\ \midrule
Epoch & 150 & 150 & 200 & 150 \\
Time(s) & 403.50 & 883.50 & 21010.00 & 4778.25 \\
\bottomrule
\end{tabular}}}
\label{table1}
\\ \hspace*{\fill} 
\\ \hspace*{\fill} \\
\centering
\caption {Parameter efficiency of embedding networks. P stands for the number of parameters, and FLOPs denotes the number of multiply-adds.}
\setlength{\tabcolsep}{3mm}{
\scalebox{0.8}{
\begin{tabular}{ccc}
\toprule
~ & ResNet-12 & DCN(Ours) \\ \midrule
P & 12.43M & 12.90M \\
FLOPs & 3523.03M & 3533.42M \\
\bottomrule
\end{tabular}}}
\label{table1}
\end{table}

Table V present the results of the evaluation of the training speed of the proposed DCN method. The training is conducted on four different datasets, with the number of epochs varying across the datasets: 200 epochs for the NWPU-RESISC45 dataset and 150 epochs for the remaining three datasets. The maximum training time is less than 6 hours.

Moreover, the parameter efficiency of our DCN is illustrated in Table VI. The comparison of the parameter quantity and calculation quantity between the embedding network in our DCN method and the original ResNet-12 reveals that the number of parameters and FLOPs of the DCN is slightly higher than that of ResNet-12. However, our proposed DCN achieves a substantial improvement over the baseline in terms of performance, demonstrating its high computational efficiency.

\section{Conclusion}
In this work, we have proposed a novel DCN approach based on the transfer learning framework for FS-RSISC, which consists of two different branches for extracting complementary features and implements the integration of supervised contrastive learning into the few-shot classification task. Specifically, the CCL branch includes a Condenser Network and a contrastive learning module, which promotes the model to extract discriminative features by exploiting the global context information. The DCL branch is designed for capturing invariant local details, which contains a Smelter Network and a contrastive learning module based on feature maps. The competitive experimental results on four popular benchmark remote sensing datasets demonstrate the effectiveness of our proposed DCN. In the future work, we will explore the semi-supervised and weakly-supervised settings to exploit the more unlabeled data and noisy data.

\bibliographystyle{IEEEtran}
\bibliography{IEEEabrv,abrvbib}

\end{document}